\documentclass{article}

\usepackage[preprint,nonatbib]{neurips_2026}

\usepackage[utf8]{inputenc}
\usepackage[T1]{fontenc}
\usepackage{hyperref}
\usepackage{url}
\usepackage{booktabs}
\usepackage{amsfonts}
\usepackage{nicefrac}
\usepackage{microtype}
\usepackage{xcolor}
\usepackage{graphicx}
\usepackage{amsmath}
\usepackage{amssymb}
\usepackage{multirow}
\usepackage{caption}
\usepackage{subcaption}
\usepackage{array}
\usepackage{enumitem}

\setlist[itemize]{leftmargin=*,topsep=2pt,itemsep=1pt}

\newcommand{\InfU}{InfiniteYou}
\newcommand{\InfuseNet}{InfuseNet}
\newcommand{\dev}{FLUX.1-dev}
\newcommand{\schnell}{FLUX.1-schnell}
\newcommand{\idsim}{ID\textsubscript{sim}}
\newcommand{\lpips}{LPIPS}

\definecolor{better}{RGB}{0,130,45}
\definecolor{worse}{RGB}{180,0,0}
\definecolor{paperblue}{RGB}{30,95,170}

\DeclareRobustCommand{\figfile}[1]{}

\newcommand{\safeincludegraphics}[2][]{%
  \IfFileExists{#2}{\includegraphics[#1]{#2}}{%
    \fbox{\parbox[c][0.22\textheight][c]{0.92\linewidth}{\centering\small Missing figure file:\\ \texttt{#2}}}%
  }%
}

\title{When Few Steps Are Enough:\\
Training-Free Acceleration of Identity-Preserved Generation}

\author{
Dongqi Zheng \\
\texttt{dqzheng1996@gmail.com}
}

\begin{document}
\maketitle

\begin{abstract}
Identity-preserved image generation is typically built on many-step diffusion backbones, making personalized generation expensive at deployment time. We show that this cost is often unnecessary for identity-conditioned FLUX generation. A frozen \InfuseNet{} identity adapter trained with \dev{} transfers directly to the distilled \schnell{} backbone without retraining. This two-line replacement---changing the backbone path and disabling classifier-free guidance---reduces latency by $5.9\times$ while improving ArcFace identity similarity by $+0.028$ and \lpips{} by $-0.016$ over the standard 28-step \dev{} baseline. To explain why this works, we analyze the denoising trajectory and find that identity fidelity enters an early effective regime, often within 4--8 steps, while later steps primarily refine visual detail, sharpness, and contrast. Adapter ablations confirm that identity formation depends on the identity adapter, while attention-stream norm probes suggest that the relative conditioning contribution decreases as sampling proceeds. Preliminary style-adapter and object-adapter sweeps on SDXL and SD1.5 show similar diminishing returns after intermediate steps. These results position distilled backbone replacement as a simple, training-free strategy for improving the efficiency--fidelity tradeoff of identity-preserved generation.
\end{abstract}

\section{Introduction}
\label{sec:intro}

Identity-preserved image generation aims to synthesize a person in new scenes, outfits, poses, and visual styles while maintaining recognizable facial identity. This capability is central to personalized content creation, virtual try-on, portrait editing, and avatar generation. Recent methods improve identity preservation by injecting face representations into pretrained text-to-image diffusion models through cross-attention adapters, residual side networks, or ControlNet-style branches~\cite{ye2023ip,wang2024instantid,guo2024pulid,jiang2025infiniteyou}. \InfU{} is a representative example: it injects ArcFace identity embeddings~\cite{deng2019arcface} into FLUX Diffusion Transformer (DiT) blocks~\cite{flux2024} through a residual identity adapter, \InfuseNet{}~\cite{jiang2025infiniteyou}.

Despite strong progress in adapter design, the sampling schedule is usually treated as a fixed engineering detail. In practice, identity-conditioned FLUX pipelines often inherit the default many-step schedule of the base backbone. For \InfU{} with \dev{}, this means a 28-step denoising trajectory that costs roughly 10 seconds per image on our hardware. The implicit assumption is that additional denoising steps improve all objectives: visual quality, text alignment, and identity fidelity.

This paper asks a deployment-oriented question: \emph{do identity-conditioned generators need the full many-step trajectory to preserve identity?} Our answer is no, at least for the FLUX-based identity setting studied here. We find that identity similarity emerges early and reaches a high-fidelity regime within a small number of steps. The exact peak is subject- and prompt-dependent---often around 4--8 steps in our stress-test sweeps---so we do not claim that four steps are universally optimal. Instead, we identify an \emph{early identity-effective window}: a region of the denoising trajectory where identity-specific facial structure has already formed, while later steps mainly sharpen and stylize the image. In this framing, \schnell{} 4-step sampling is a practical deployment point inside or near this early window, not a universal optimum.

This observation motivates an extremely simple method. We replace the 28-step \dev{} backbone in \InfU{} with the distilled 4-step \schnell{} backbone while keeping \InfuseNet{} frozen. The replacement requires only two configuration changes: the base model path is changed from \dev{} to \schnell{}, and classifier-free guidance is disabled as expected for \schnell{}. No identity adapter weights, face encoder weights, prompts, or conditioning scales are retrained or tuned.

Empirically, this training-free replacement improves the efficiency--fidelity tradeoff. On 28 FFHQ identities and three prompts per identity, \schnell{} 4-step achieves $5.9\times$ lower latency, improves ArcFace similarity from 0.5872 to 0.6150, and improves \lpips{} from 0.7253 to 0.7097 relative to the 28-step \dev{} baseline. A \dev{} step sweep and mechanistic probes explain the result: identity similarity rises rapidly, the adapter identity lift is large after a few steps, visual sharpness and contrast continue increasing after identity saturation, and the relative attention-stream norm decreases through denoising. Thus, \schnell{} is not better because step 4 is a universal identity optimum; it is useful because a distilled few-step trajectory lands in the early identity-effective region while avoiding expensive late-stage refinement.

\paragraph{Contributions.}
\begin{itemize}
  \item \textbf{Training-free distilled backbone replacement.} We show that a frozen \InfuseNet{} identity adapter trained with \dev{} transfers directly to \schnell{} without retraining.
  \item \textbf{Improved efficiency--fidelity tradeoff.} The 4-step distilled backbone achieves $5.9\times$ lower latency while improving identity similarity by $+0.028$ and \lpips{} by $-0.016$.
  \item \textbf{Early identity-effective window.} We show that identity fidelity rises rapidly and reaches a stable high-fidelity regime within a few denoising steps, making the full 28-step trajectory inefficient for identity preservation.
  \item \textbf{Mechanistic analysis.} Adapter ablation, visual refinement probes, and attention-stream norm probes suggest that identity forms early, while later denoising primarily improves visual detail and reduces the relative conditioning-stream contribution.
  \item \textbf{Broader sanity checks.} Preliminary SDXL style-adapter and SD1.5 object-adapter sweeps show similar diminishing returns before the default endpoint, suggesting that effective sampling windows may be common in adapter-conditioned generation.
\end{itemize}

\section{Background}
\label{sec:background}

\subsection{Identity-Preserved Generation}

Adapter-based personalization methods condition text-to-image diffusion models on reference images or identity embeddings. IP-Adapter injects image features using decoupled cross-attention~\cite{ye2023ip}. InstantID and PuLID improve identity alignment with stronger face-specific representations and objectives~\cite{wang2024instantid,guo2024pulid}. \InfU{} extends identity preservation to FLUX DiT backbones by using \InfuseNet{} to inject ArcFace embeddings as residual signals across transformer blocks~\cite{jiang2025infiniteyou,deng2019arcface,flux2024}.

In this paper, we focus on \InfU{} because it gives a natural testbed for training-free backbone replacement. The identity pathway is separate from the FLUX backbone, but is inserted into the backbone hidden-state trajectory. This makes it possible, but not guaranteed, that a frozen adapter trained with \dev{} can still operate on the distilled \schnell{} representation space.

\subsection{FLUX and Distilled Few-Step Sampling}

FLUX uses rectified flow~\cite{lipman2022flow,liu2022rectified,flux2024}, which interpolates between data $x_0$ and noise $\varepsilon$:
\begin{equation}
  x_t = (1-t)x_0 + t\varepsilon, \qquad \varepsilon \sim \mathcal{N}(0,I),
  \label{eq:flow}
\end{equation}
with velocity target $v^* = x_0 - \varepsilon$. Rectified-flow and consistency-distillation objectives enable few-step generation more naturally than classical many-step DDPM/DDIM sampling~\cite{ho2020denoising,song2021ddim,song2023consistency,salimans2022progressive}. \schnell{} is a distilled FLUX backbone intended for 4-step sampling and is typically used with classifier-free guidance disabled.

\subsection{Residual Identity Conditioning}

We write the identity adapter update at transformer block $\ell$ as
\begin{equation}
  h'_{\ell} = h_{\ell} + \alpha \cdot \InfuseNet_{\ell}(e_{\mathrm{id}}),
  \label{eq:adapter}
\end{equation}
where $e_{\mathrm{id}}$ is an ArcFace embedding and $\alpha$ is the identity-conditioning scale. This residual formulation is important: the adapter nudges the backbone representation rather than replacing it. If distillation preserves enough hidden-state geometry, the same residual adapter can remain effective on a distilled backbone.

\section{Training-Free Distilled Backbone Replacement}
\label{sec:method}

Our method is intentionally minimal. Starting from an \InfU{} pipeline using \dev{} and frozen \InfuseNet{} weights, we change only:
\begin{center}
\begin{tabular}{lll}
\toprule
\textbf{Configuration} & \textbf{Baseline} & \textbf{Ours} \\
\midrule
Backbone path & \texttt{FLUX.1-dev} & \texttt{FLUX.1-schnell} \\
Inference steps & 28 & 4 \\
Guidance scale & 3.5 & 0.0 \\
\bottomrule
\end{tabular}
\end{center}

\begin{figure}[t]
  \centering
  \safeincludegraphics[width=0.98\linewidth]{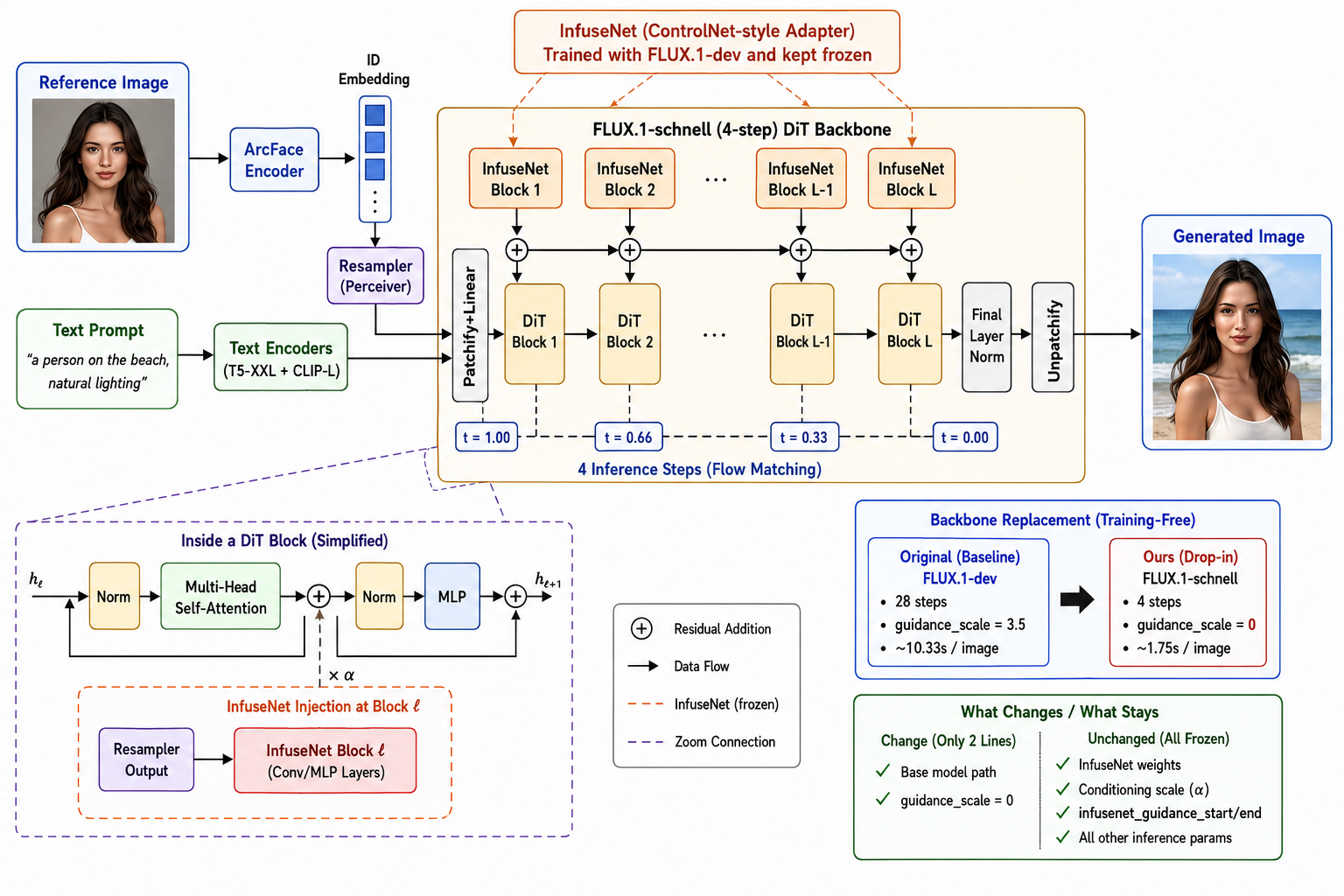}
  \caption{\textbf{Training-free distilled backbone replacement.} We keep the ArcFace encoder and the \InfuseNet{} identity adapter frozen, and only replace the base DiT backbone from \dev{} to \schnell{} while setting guidance scale to zero. The identity adapter remains a residual injection path into the DiT blocks; no identity-specific weights are retrained.}
  \label{fig:architecture}
\end{figure}

Figure~\ref{fig:architecture} summarizes the training-free replacement. The architecture and identity pathway remain unchanged; only the base backbone and guidance-scale setting are modified at inference time.

All \InfuseNet{} weights, identity encoder weights, prompts, resolution, and identity-conditioning scales are kept fixed. We do not fine-tune, distill, LoRA-adapt~\cite{hu2022lora}, or recalibrate the identity adapter.

\paragraph{Why this should work.}
We hypothesize that identity-conditioned generation has an early identity-effective window. In early denoising, the model forms coarse facial structure and identity-specific geometry; after this point, additional steps refine texture, contrast, lighting, background, and style. A distilled few-step model can be effective if it reaches the identity-effective region and has been trained to produce visually stable outputs at that low step count~\cite{salimans2022progressive,song2023consistency}. In contrast, naively truncating a non-distilled backbone can be diagnostically useful but may not be the best deployment strategy, because the original backbone was not optimized for high-quality few-step sampling. Appendix Figure~\ref{fig:theory_fit} summarizes this early-window hypothesis.

\section{Experimental Setup}
\label{sec:setup}

\paragraph{Main identity evaluation.}
We evaluate on 28 adult FFHQ identities~\cite{karras2019ffhq} resized to $512\times512$, filtered for a single visible face and high ArcFace~\cite{deng2019arcface} detection confidence. Each identity is evaluated on three prompts: portrait, beach, and professional suit, giving 84 samples per method.

\paragraph{Mechanistic step sweep.}
To understand why few steps are sufficient, we run a \dev{} step sweep on an anime-style stress-test prompt, \emph{``a person in anime illustration style''}. We use steps $T \in \{1,2,4,6,8,12,16,20,24,28\}$ and 10 identities. This prompt intentionally creates a strong style shift, making the competition between identity preservation and visual stylization more visible.

\paragraph{Baselines.}
The main baseline is \InfU{} with \dev{} for 28 steps. We also include a \dev{} 4-step diagnostic on a smaller subset to isolate early stopping from distillation. Our method is \InfU{} with \schnell{} for 4 steps and frozen \InfuseNet{} weights.

\paragraph{Metrics.}
We report ArcFace cosine similarity~\cite{deng2019arcface} (\idsim{}, higher is better), \lpips{} with a VGG backbone~\cite{zhang2018unreasonable} (lower is better), and wall-clock latency per image. For mechanistic analysis, we additionally measure face detection confidence, face area ratio, image sharpness via Laplacian variance, image contrast, adapter identity lift $\idsim(\alpha{=}1)-\idsim(\alpha{=}0.25)$, and attention-stream norm ratios from FLUX attention outputs. We do not interpret the stream-ratio probe as literal attention probability; it is a representation-norm diagnostic.

\paragraph{Preliminary non-identity adapter sweeps.}
In the appendix, we also report preliminary SDXL style-adapter and SD1.5 object-adapter step sweeps using CLIP-image and DINOv2-style perceptual signals~\cite{oquab2023dinov2}. These are used only as sanity checks for diminishing returns and are not part of the main identity-preserved generation claim.

\section{Main Results: Efficient Identity-Preserved Generation}
\label{sec:main_results}

\subsection{Distilled Backbone Replacement Improves the Efficiency--Fidelity Tradeoff}

Table~\ref{tab:main} shows the main quantitative result. Replacing \dev{} with \schnell{} reduces latency from 10.29s to 1.73s, a $5.9\times$ speedup. More importantly, the speedup does not come at the cost of identity fidelity: \idsim{} improves from 0.5872 to 0.6150, and \lpips{} improves from 0.7253 to 0.7097.

\begin{table}[t]
\centering
\caption{\textbf{Main identity-preservation results} on 28 identities $\times$ 3 prompts. \schnell{} improves latency, identity similarity, and perceptual distance without retraining the identity adapter. The \dev{} 4-step row is a diagnostic subset and should not be interpreted as a full baseline.}
\label{tab:main}
\small
\setlength{\tabcolsep}{6pt}
\begin{tabular}{lrrrrr}
\toprule
\textbf{Method} & \textbf{Steps} & \textbf{\idsim{} $\uparrow$} & \textbf{\lpips{} $\downarrow$} & \textbf{Latency $\downarrow$} & \textbf{Speedup} \\
\midrule
\dev{} 28-step baseline & 28 & 0.5872 & 0.7253 & 10.29s & 1.0$\times$ \\
\dev{} 4-step diagnostic$^{\dagger}$ & 4 & 0.6085 & -- & 1.73s & 5.9$\times$ \\
\textbf{\schnell{} 4-step (ours)} & 4 & \textbf{0.6150} & \textbf{0.7097} & \textbf{1.73s} & \textbf{5.9$\times$} \\
\midrule
$\Delta$ vs. baseline & -- & \textcolor{better}{$+0.0278$} & \textcolor{better}{$-0.0156$} & \textcolor{better}{$-8.56$s} & -- \\
\bottomrule
\end{tabular}
\vspace{2pt}

\footnotesize{$^{\dagger}$Measured on a smaller 10-face portrait subset; included to separate early stopping from distilled-backbone deployment.}
\end{table}

\subsection{Per-Prompt Consistency}

Table~\ref{tab:prompt} breaks down identity similarity by prompt type. Appendix Figure~\ref{fig:results_bars} shows the per-identity distribution. The gains are consistent across portrait, beach, and suit prompts, suggesting that the improvement is not driven by a single scene type.

\begin{table}[t]
\centering
\caption{\textbf{\idsim{} by prompt type}. Gains are consistent across all three evaluation prompts.}
\label{tab:prompt}
\small
\setlength{\tabcolsep}{12pt}
\begin{tabular}{lccc}
\toprule
\textbf{Prompt} & \textbf{\dev{} 28-step} & \textbf{\schnell{} 4-step} & \textbf{$\Delta$} \\
\midrule
Portrait & 0.5752 & \textbf{0.6080} & \textcolor{better}{$+0.0328$} \\
Beach & 0.5908 & \textbf{0.6178} & \textcolor{better}{$+0.0270$} \\
Suit & 0.5956 & \textbf{0.6193} & \textcolor{better}{$+0.0237$} \\
\midrule
Average & 0.5872 & \textbf{0.6150} & \textcolor{better}{$+0.0278$} \\
\bottomrule
\end{tabular}
\end{table}

\subsection{Qualitative Comparison}

\begin{figure}[t]
  \centering
  \safeincludegraphics[width=0.95\linewidth]{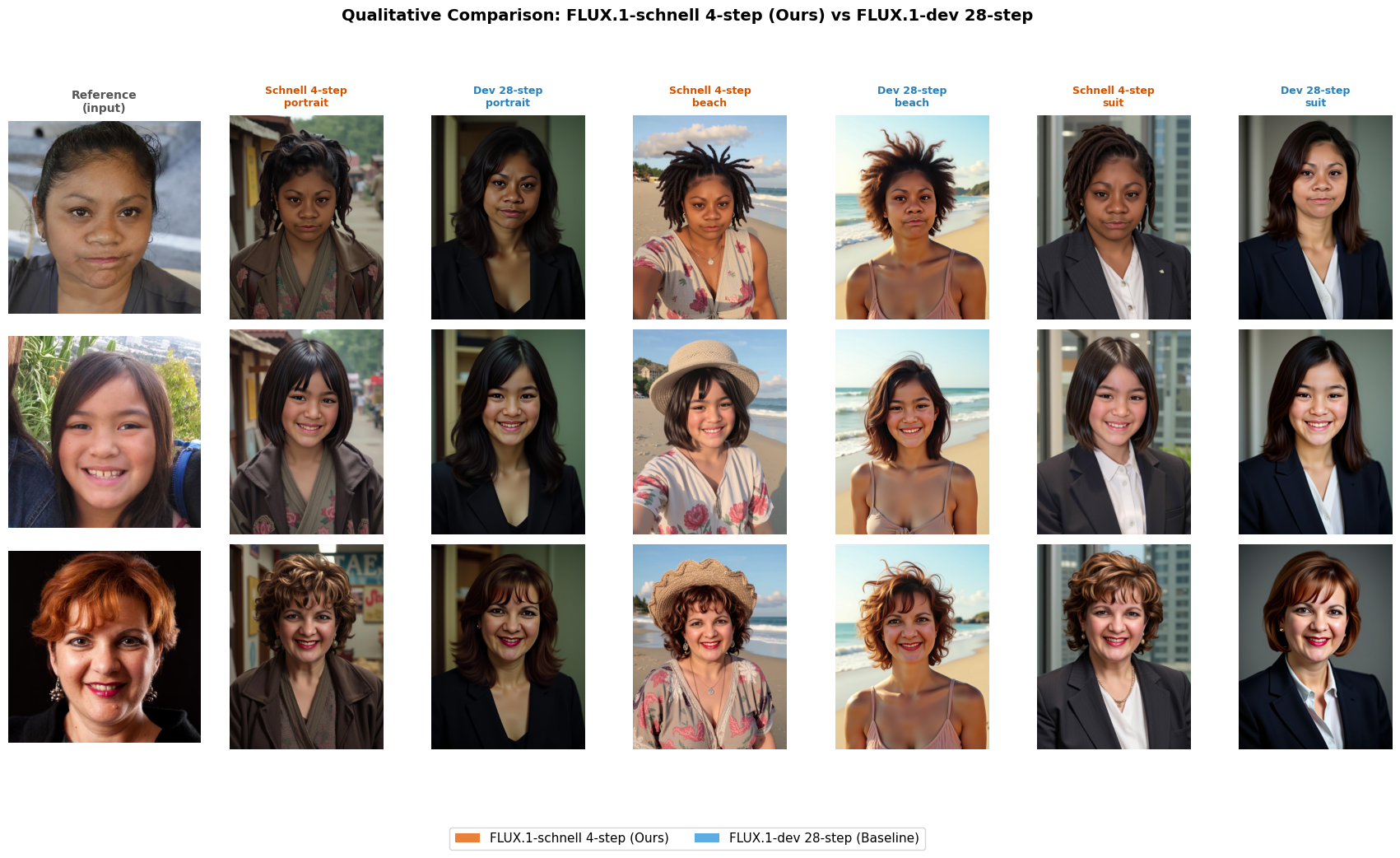}
  \caption{\textbf{Qualitative comparison.} For each reference identity, \schnell{} 4-step outputs are compared with \dev{} 28-step outputs under portrait, beach, and suit prompts. The comparison illustrates that the distilled backbone preserves recognizable identity while reducing latency. \figfile{qualitative.png}}
  \label{fig:qual}
\end{figure}

Figure~\ref{fig:qual} compares representative generated images. The distilled backbone preserves the coarse facial structure and identity attributes while reducing generation time by nearly $6\times$. We emphasize that the method does not rely on a new identity loss or adapter training; it is a deployment-time backbone substitution. A per-identity quantitative breakdown is deferred to Appendix Figure~\ref{fig:results_bars}.

\section{Why Few Steps Are Enough}
\label{sec:why}

The main result raises a natural question: why can a 4-step distilled backbone match or exceed a 28-step baseline for identity preservation? We answer this through a step sweep and mechanistic probes. The key message is not that four steps are always optimal, but that identity enters an early effective window before the default endpoint.

\subsection{Identity Emerges in an Early Denoising Window}

In the \dev{} step sweep, identity similarity rises rapidly from near zero to a high-fidelity regime within the first few steps (Table~\ref{tab:mechanism_summary}). The mean \idsim{} is 0.505 at 4 steps, 0.572 at 6 steps, and 0.615 at 8 steps, after which it saturates or slightly declines. By 28 steps, \idsim{} is 0.589. Thus, later steps do not provide proportional identity gains, even though they continue to refine visual details.

\begin{figure*}[t]
  \centering
  \safeincludegraphics[width=0.98\textwidth]{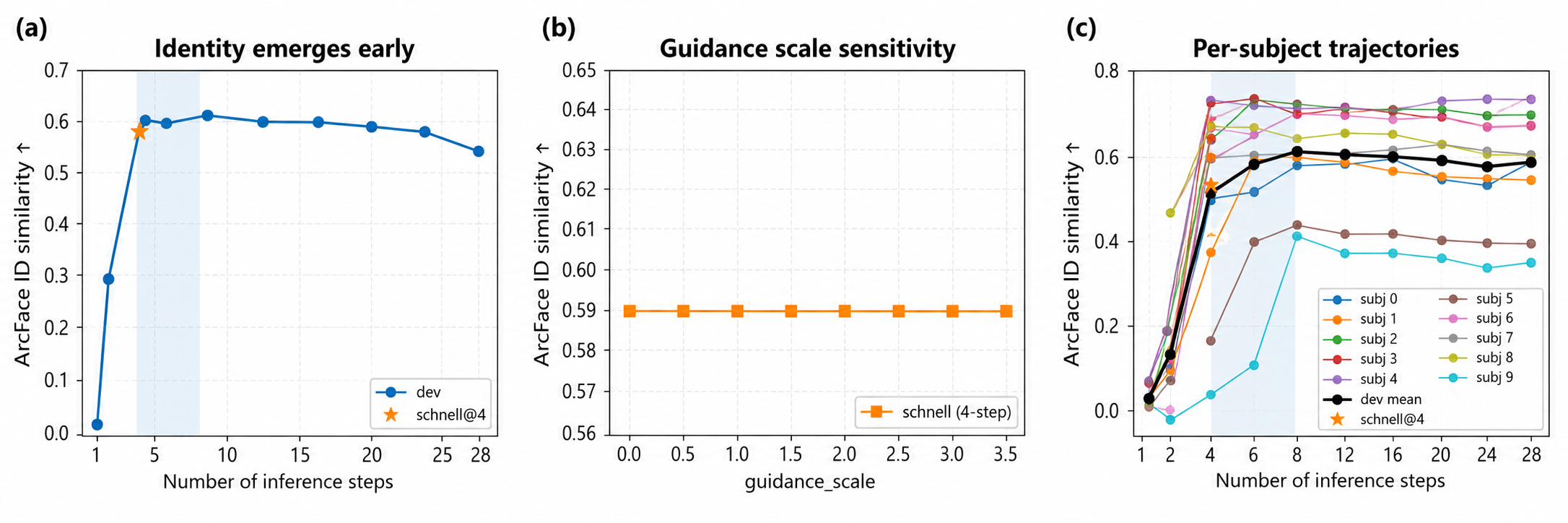}
  \caption{\textbf{Identity emerges in an early effective window.} (a) \dev{} identity similarity rises quickly and enters a high-fidelity region around 4--8 steps, while the \schnell{} 4-step deployment point lands near this region. (b) \schnell{} identity similarity is nearly insensitive to the tested guidance-scale values. (c) Individual subjects have different peak steps, so we use an early-window interpretation rather than claiming a universal 4-step optimum.}
  \label{fig:early_window_combined}
\end{figure*}

\subsection{Per-Subject Variation}

The identity-effective window is not identical for all subjects. Some identities peak around 4 steps, while others peak around 6 or 8 steps. This subject dependence is why we avoid the stronger claim that 4 steps is universally optimal. The practical observation is weaker but more robust: most identity information is already present before the 28-step endpoint.

\subsection{Later Steps Refine Visual Quality More Than Identity}

Visual probes explain why the 28-step outputs may still look more refined even when identity has saturated. Face detection confidence increases from 0.585 at step 1 to 0.858 at step 28; image sharpness increases from 19.6 to 445.2; and image contrast increases from 48.3 to 71.8. In contrast, identity similarity peaks earlier and does not increase proportionally. This supports a separation between early identity formation and late visual refinement.

\begin{figure*}[t]
  \centering
  \safeincludegraphics[width=0.98\textwidth]{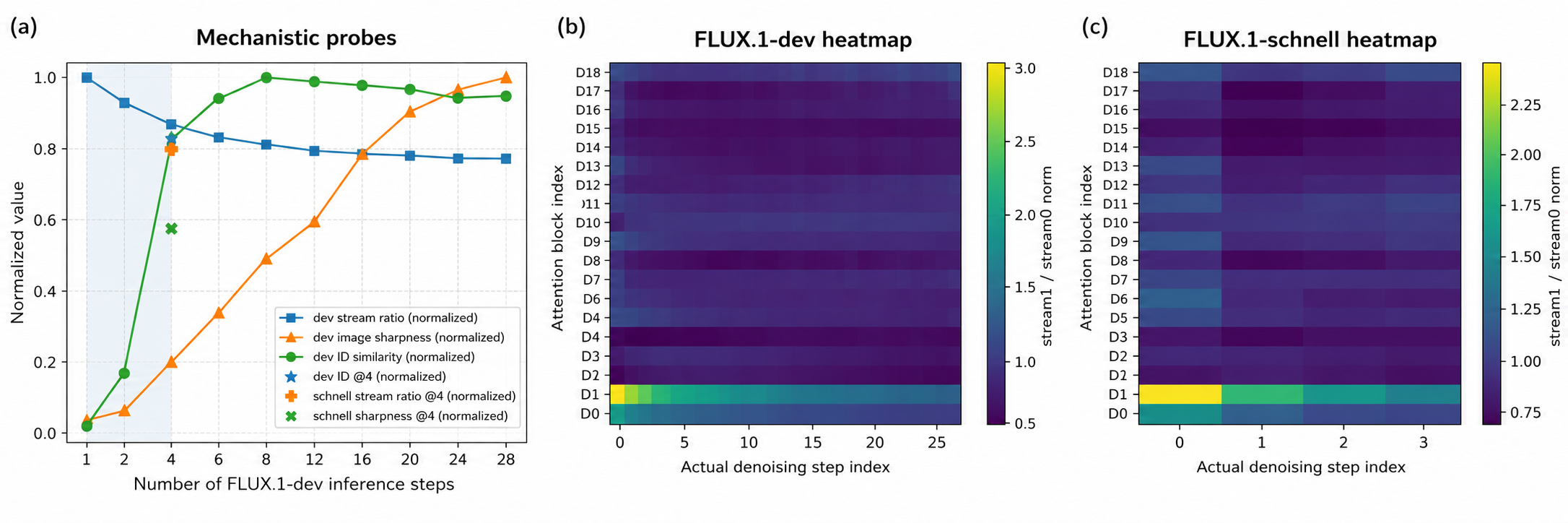}
  \caption{\textbf{Mechanistic probes.} (a) Identity similarity saturates early, while image sharpness continues increasing; the attention-stream norm ratio decreases over denoising steps. (b,c) Block-wise heatmaps show the same representation-norm diagnostic for \dev{} and \schnell{}. We interpret these ratios conservatively as stream-output norm measurements, not literal attention probabilities.}
  \label{fig:mechanism_combined}
\end{figure*}

\subsection{Adapter Ablation Confirms Identity Conditioning Is Necessary}

To test whether identity formation actually comes from the adapter, we compare the main identity scale $\alpha=1.0$ with a weak-adapter setting $\alpha=0.25$. The weak-adapter similarity remains low, around 0.05--0.08 across most steps, while the main adapter reaches 0.58--0.62. The adapter identity lift rises quickly: 0.448 at 4 steps, 0.519 at 6 steps, 0.561 at 8 steps, and remains large through 28 steps. This confirms that the adapter is necessary for identity preservation, while also showing that the useful identity contribution appears early.

\subsection{Attention-Stream Probes Suggest Late Conditioning Dilution}

We also probe FLUX attention outputs by measuring the relative norm of the secondary stream to the primary stream. This ratio decreases monotonically from 1.040 at step 1 to 0.811 at step 28. Because this is a norm ratio, not an attention probability, we interpret it conservatively: it suggests that the relative contribution of the conditioning-related stream becomes smaller as denoising proceeds. Combined with the visual probes, this supports an efficiency interpretation: late denoising spends compute on visual refinement while providing diminishing returns for identity.

\begin{table}[t]
\centering
\caption{\textbf{Mechanistic step sweep} on the anime-style stress-test prompt. Identity similarity enters a high-fidelity window around 4--8 steps, while visual refinement metrics continue improving through later steps.}
\label{tab:mechanism_summary}
\scriptsize
\setlength{\tabcolsep}{4pt}
\begin{tabular}{rrrrrrr}
\toprule
\textbf{Steps} & \textbf{\idsim{} $\alpha{=}1$} & \textbf{\idsim{} $\alpha{=}0.25$} & \textbf{Lift} & \textbf{Stream ratio} & \textbf{Face det.} & \textbf{Sharpness} \\
\midrule
1  & 0.015 & --    & --    & 1.040 & 0.585 & 19.6 \\
2  & 0.105 & 0.036 & 0.066 & 0.970 & 0.693 & 31.1 \\
4  & 0.505 & 0.057 & 0.448 & 0.914 & 0.801 & 91.8 \\
6  & 0.572 & 0.053 & 0.519 & 0.882 & 0.813 & 151.0 \\
8  & 0.615 & 0.055 & 0.561 & 0.865 & 0.796 & 210.2 \\
12 & 0.609 & 0.051 & 0.548 & 0.844 & 0.832 & 263.6 \\
16 & 0.605 & 0.063 & 0.555 & 0.831 & 0.836 & 347.8 \\
20 & 0.597 & 0.058 & 0.604 & 0.822 & 0.848 & 403.3 \\
24 & 0.580 & 0.057 & 0.581 & 0.814 & 0.858 & 427.5 \\
28 & 0.589 & 0.081 & 0.555 & 0.811 & 0.858 & 445.2 \\
\bottomrule
\end{tabular}
\end{table}

\subsection{Ablation Studies}

\begin{figure}[t]
  \centering
  \safeincludegraphics[width=0.95\linewidth]{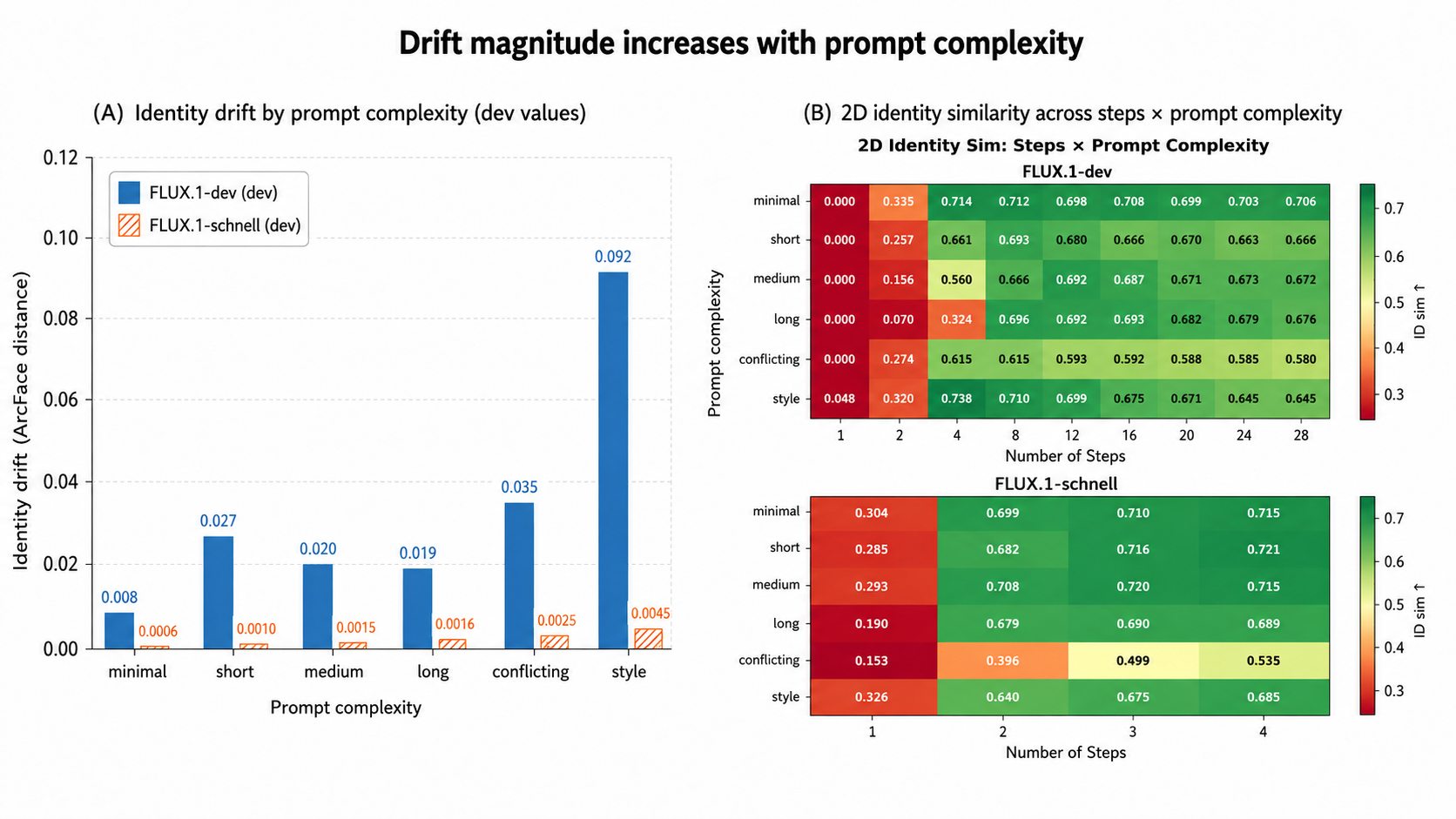
}
  \caption{\textbf{Prompt-complexity ablation.} Drift magnitude is small for simple prompts and larger for style-conflicting prompts. The paired heatmaps summarize identity similarity across steps and prompt complexity for \dev{} and \schnell{}.}
  \label{fig:prompt_complexity}
\end{figure}

\begin{figure}[t]
  \centering
  \safeincludegraphics[width=0.95\linewidth]{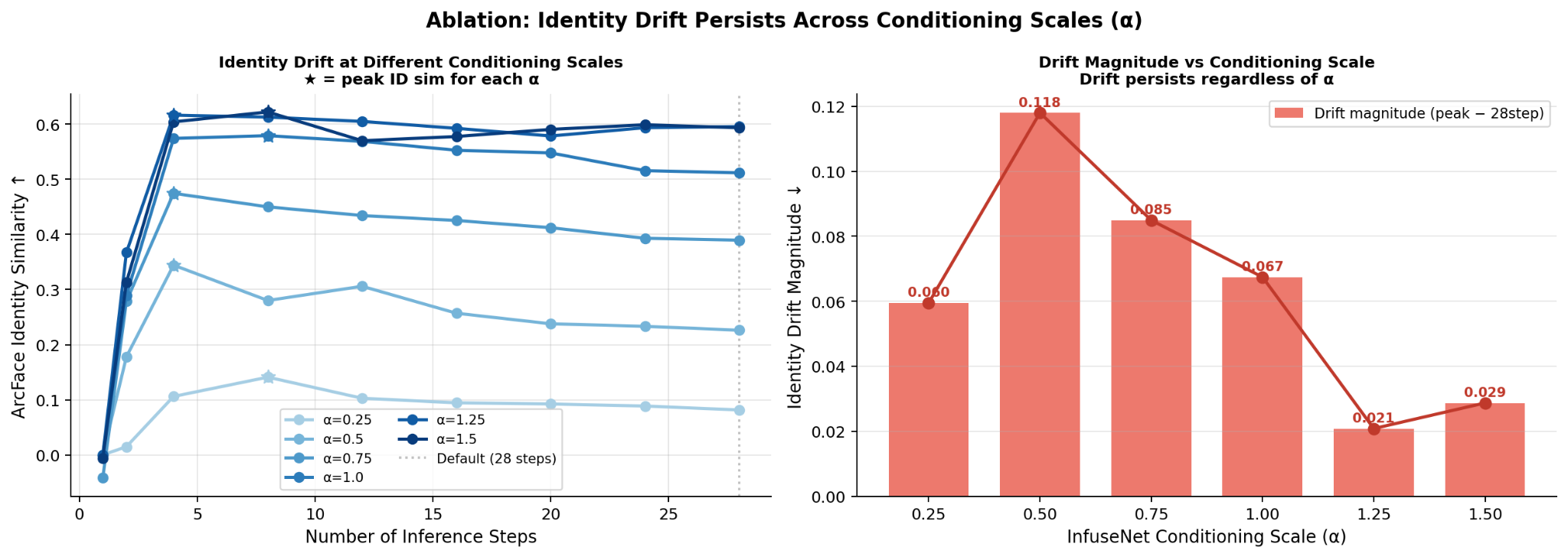}
  \caption{\textbf{Conditioning-scale ablation.} Identity drift/saturation appears across adapter scales. Scaling changes the absolute similarity and peak location, but does not make the late default endpoint necessary for identity preservation.}
  \label{fig:alpha_ablation}
\end{figure}

Figure~\ref{fig:prompt_complexity} studies prompt complexity and shows that drift becomes more visible under stronger prompt or style conflict. Figure~\ref{fig:alpha_ablation} varies the adapter conditioning scale and shows that the early-window behavior persists across a broad range of \(\alpha\) values, even though the absolute identity level changes.

\section{Preliminary Evidence Beyond FLUX}
\label{sec:beyond_flux}

Our main claim is about FLUX-based identity generation because \dev{} and \schnell{} form a natural original/distilled backbone pair. We do not claim universal adapter drift across all models. However, preliminary step sweeps on other adapter-conditioned pipelines show a related phenomenon: adapter fidelity often reaches a useful regime before the default endpoint.

\paragraph{SDXL style adapter.}
For style conditioning on SDXL, CLIP image similarity improves rapidly from 0.732 at 5 steps to 0.800 at 20 steps, while the 50-step endpoint is 0.806. Thus, 20 steps reach 99.2\% of the 50-step style fidelity with 43.0\% of the latency.

\paragraph{SD1.5 object adapter.}
For object conditioning on SD1.5, DINOv2 similarity improves from 0.218 at 5 steps to 0.455 at 20 steps, while the 50-step endpoint is 0.473. Thus, 20 steps reach 96.2\% of the 50-step object fidelity with 44.4\% of the latency.

These results should be interpreted as sanity checks rather than full cross-backbone generalization. They support the broader hypothesis that adapter-conditioned generation has task-specific effective sampling windows, but a complete study would need to control for sampler, scheduler, adapter architecture, metric, and backbone family.

\section{Related Work}
\label{sec:related}

\paragraph{Identity-preserved generation.}
IP-Adapter injects image features into text-to-image diffusion models through decoupled cross-attention~\cite{ye2023ip}. InstantID and PuLID improve zero-shot identity preservation using stronger face encoders and contrastive alignment~\cite{wang2024instantid,guo2024pulid}. \InfU{} extends identity-preserved generation to FLUX backbones using \InfuseNet{}~\cite{jiang2025infiniteyou}. These works primarily optimize adapter design; we study the deployment-time backbone and sampling schedule.

\paragraph{Diffusion acceleration.}
DDPM requires many denoising steps~\cite{ho2020denoising}. DDIM, progressive distillation, consistency models, and rectified flow reduce sampling cost~\cite{song2021ddim,salimans2022progressive,song2023consistency,lipman2022flow,liu2022rectified}. Our work is complementary: rather than proposing a new sampler, we show that an existing distilled backbone can be used as a training-free replacement for identity-preserved generation.

\paragraph{Adapter transfer.}
LoRA and model soups show that learned modifications can transfer or combine across related checkpoints~\cite{hu2022lora,wortsman2022model}. We study a different form of transfer: a residual identity adapter trained with one backbone is deployed zero-shot on a distilled counterpart.

\paragraph{Step-count effects.}
Prior diffusion work often studies the tradeoff between step count and visual quality. We focus on identity fidelity and show that identity preservation can saturate earlier than visual refinement, making default many-step schedules inefficient for personalized generation.

\section{Limitations and Responsible Use}
\label{sec:limitations}

\paragraph{Limitations.}
First, the optimal identity step is subject- and prompt-dependent; we therefore claim an early identity-effective window rather than a universal 4-step optimum. Second, ArcFace similarity~\cite{deng2019arcface} is an imperfect identity metric, especially under strong stylization. Human preference studies and face-region perceptual metrics would provide additional evidence. Third, our mechanism probes are diagnostic, not definitive causal proof. Attention-stream norm ratios should not be interpreted as attention probabilities. Fourth, the strongest result relies on FLUX family compatibility between \dev{} and \schnell{}; other backbones may not support drop-in adapter transfer. Finally, our main identity evaluation uses 28 identities and three prompts, which is sufficient for the present deployment study but not exhaustive across demographics, poses, and artistic styles.

\paragraph{Responsible use.}
Identity-preserved face generation can be misused for impersonation. Deployment should require subject consent, provenance tracking, watermarking or disclosure where appropriate, and safeguards against generating deceptive images of real people. Our contribution is a technical analysis of sampling efficiency and identity fidelity, not an endorsement of unrestricted face synthesis.

\section{Conclusion}
\label{sec:conclusion}

We show that identity-preserved FLUX generation can be accelerated without retraining by replacing the 28-step \dev{} backbone with the distilled 4-step \schnell{} backbone while keeping \InfuseNet{} frozen. The method reduces latency by $5.9\times$ and improves both identity similarity and \lpips{} in our evaluation. Mechanistic analysis shows why this is possible: identity information emerges early, while later denoising mainly refines visual details and provides diminishing returns for identity preservation. More broadly, our results suggest that sampling budget should be treated as a first-class deployment parameter for adapter-conditioned generation.


\appendix

\section{Conceptual Early-Window Model}
\label{app:theory_fit}

\begin{figure}[h]
  \centering
  \safeincludegraphics[width=0.68\linewidth]{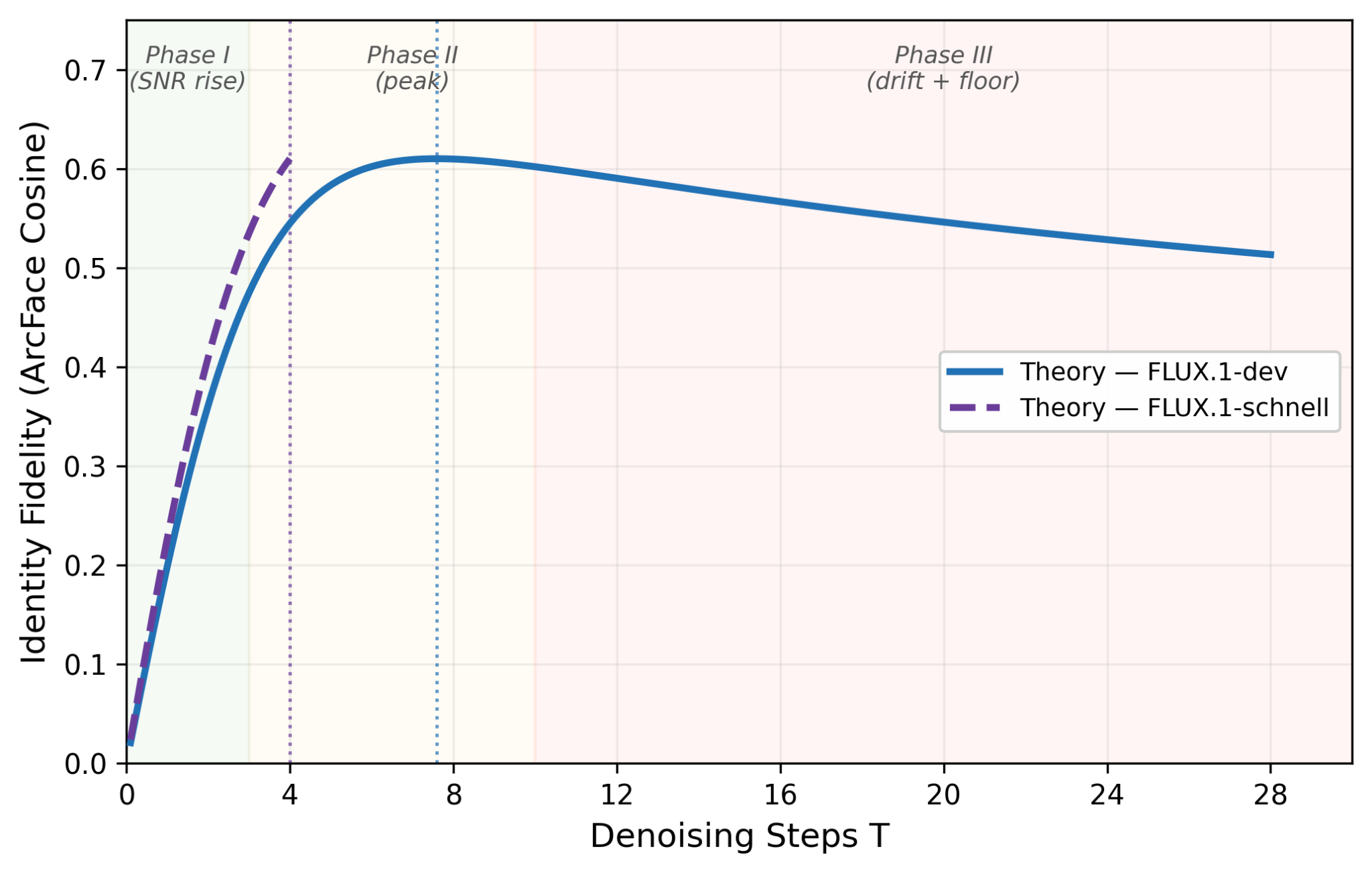}
  \caption{\textbf{Conceptual identity-fidelity trajectory.} Identity fidelity first rises as face structure emerges, enters an early effective window, and then provides diminishing identity returns while later steps mainly refine visual detail. The distilled \schnell{} trajectory is interpreted as a short deployment path into this early useful region, not as proof that exactly four steps is universally optimal.}
  \label{fig:theory_fit}
\end{figure}

Figure~\ref{fig:theory_fit} illustrates the qualitative early-window interpretation used throughout the paper. We keep this conceptual plot in the appendix because the main paper emphasizes measured deployment results and mechanistic probes.

\section{Additional Quantitative Results}
\label{app:additional_results}

\subsection{Per-Identity Breakdown on the Main Evaluation}

\begin{figure}[h]
  \centering
  \safeincludegraphics[width=0.98\linewidth]{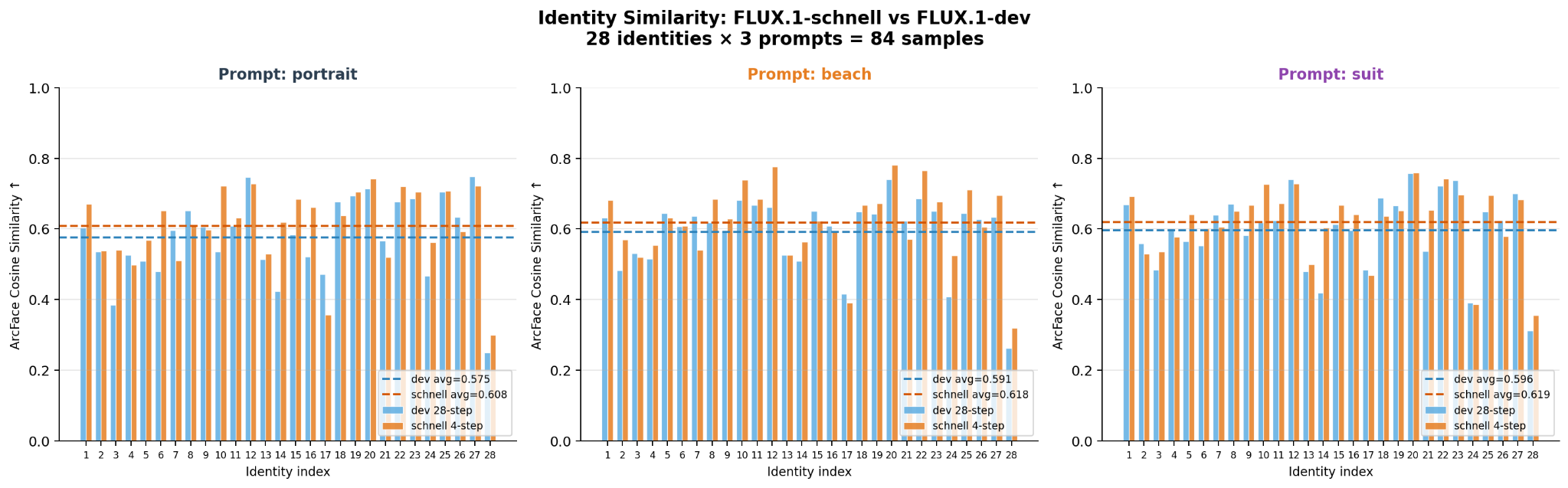}
  \caption{\textbf{Per-identity identity similarity on the main evaluation.} Across portrait, beach, and suit prompts, \schnell{} 4-step generally matches or exceeds the 28-step \dev{} baseline while using substantially fewer denoising steps. Dashed lines mark per-prompt averages.}
  \label{fig:results_bars}
\end{figure}

\subsection{SDXL Style-Adapter Sweep}

Table~\ref{tab:sdxl_style} reports the preliminary SDXL style-adapter sweep. The result is a diminishing-return pattern rather than a strong drift pattern: fidelity improves quickly and then saturates.

\begin{table}[h]
\centering
\caption{\textbf{Preliminary SDXL style-adapter sweep.} CLIP image similarity reaches 99.2\% of the 50-step endpoint by 20 steps, using 43.0\% of the latency.}
\label{tab:sdxl_style}
\small
\setlength{\tabcolsep}{7pt}
\begin{tabular}{rrrrr}
\toprule
\textbf{Steps} & \textbf{CLIP-I $\uparrow$} & \textbf{\lpips{} $\downarrow$} & \textbf{Latency} & \textbf{CLIP-I / 50-step} \\
\midrule
5  & 0.7319 & 0.6753 & 0.90s & 90.8\% \\
10 & 0.7863 & 0.6605 & 1.47s & 97.5\% \\
15 & 0.7962 & 0.6557 & 2.05s & 98.8\% \\
20 & 0.7997 & 0.6542 & 2.62s & 99.2\% \\
30 & 0.8030 & 0.6534 & 3.78s & 99.6\% \\
40 & 0.8053 & 0.6541 & 4.92s & 99.9\% \\
50 & 0.8063 & 0.6542 & 6.09s & 100.0\% \\
\bottomrule
\end{tabular}
\end{table}

\subsection{SD1.5 Object-Adapter Sweep}

Table~\ref{tab:sd15_object} reports the preliminary SD1.5 object-adapter sweep. DINOv2 similarity reaches 96.2\% of the 50-step endpoint by 20 steps, with 44.4\% of the latency.

\begin{table}[h]
\centering
\caption{\textbf{Preliminary SD1.5 object-adapter sweep.} Object fidelity reaches an effective region well before the 50-step endpoint.}
\label{tab:sd15_object}
\small
\setlength{\tabcolsep}{6pt}
\begin{tabular}{rrrrrr}
\toprule
\textbf{Steps} & \textbf{CLIP-I $\uparrow$} & \textbf{DINO $\uparrow$} & \textbf{\lpips{} $\downarrow$} & \textbf{Latency} & \textbf{DINO / 50-step} \\
\midrule
5  & 0.5521 & 0.2177 & 0.7238 & 0.37s & 46.0\% \\
10 & 0.5900 & 0.3021 & 0.7090 & 0.58s & 63.8\% \\
15 & 0.6655 & 0.4125 & 0.7018 & 0.79s & 87.1\% \\
20 & 0.6963 & 0.4554 & 0.6943 & 1.00s & 96.2\% \\
30 & 0.6974 & 0.4546 & 0.6920 & 1.42s & 96.0\% \\
50 & 0.7051 & 0.4734 & 0.6872 & 2.26s & 100.0\% \\
\bottomrule
\end{tabular}
\end{table}
\begin{figure}[!b]
  \centering
  \safeincludegraphics[width=0.9\linewidth]{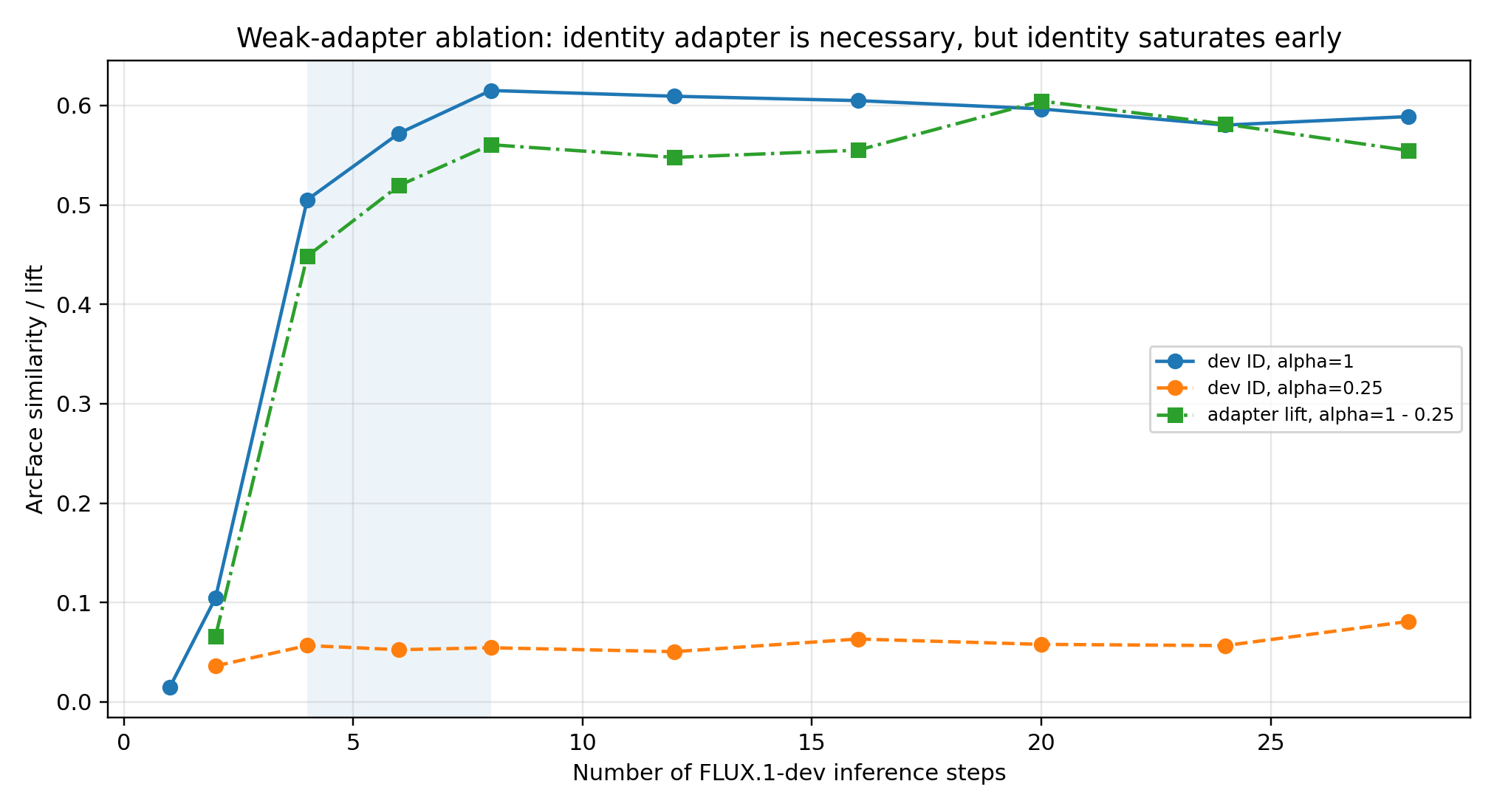}
  \caption{\textbf{Weak-adapter ablation.} Reducing the adapter scale to \(\alpha=0.25\) keeps identity similarity low, while the full adapter reaches high identity similarity after a few steps. The lift rises quickly and then largely saturates, supporting the early-effective-window interpretation. \figfile{dev\_weak\_adapter\_lift\_vs\_steps.png}}
  \label{fig:adapter_lift}
\end{figure}

\section{Mechanistic Probe Details}
\label{app:probe_details}

\paragraph{Adapter identity lift.}
We define adapter identity lift at step count $T$ as
\begin{equation}
  \Delta_{\mathrm{id}}(T) = \idsim(T, \alpha=1.0) - \idsim(T, \alpha=0.25).
\end{equation}
We use $\alpha=0.25$ rather than $\alpha=0$ in the final probe because fully disabling the adapter can make face detection unstable under strongly stylized prompts. The weak-adapter baseline keeps images face-like while reducing identity anchoring.

\paragraph{Attention-stream norm ratio.}
For FLUX attention modules whose forward output exposes two tensor streams, we compute
\begin{equation}
  R(T,\ell) = \frac{\|s_1(T,\ell)\|_2}{\|s_0(T,\ell)\|_2 + \epsilon},
\end{equation}
where $s_0$ and $s_1$ denote the two output streams at denoising step $T$ and block $\ell$. We average $R$ over blocks and subjects. This is a representation-norm diagnostic and should not be interpreted as literal attention mass.

\end{document}